\documentclass[final]{cvpr}

\usepackage{times}
\usepackage{epsfig}
\usepackage{graphicx}
\usepackage{amsmath}
\usepackage{amssymb}

\usepackage{dsfont}
\usepackage{array}
\usepackage{booktabs}

\usepackage[pagebackref=true,breaklinks=true,colorlinks,bookmarks=false]{hyperref}

\def\cvprPaperID{376} 
\def\confYear{CVPR 2021}

\begin{document}
	
\title{Distilling Object Detectors via Decoupled Features}

\author{Jianyuan Guo$^{1,2}$, Kai Han$^{1}$, Yunhe Wang$^{1}$\thanks{Corresponding author.}, Han Wu$^{2}$, Xinghao Chen$^{1}$, Chunjing Xu$^{1}$, Chang Xu$^{2*}$ \\
	\normalsize$^1$ Noah's Ark Lab, Huawei Technologies.
	\\
	\normalsize$^2$ School of Computer Science, Faculty of Engineering, University of Sydney.\\
	\small\texttt{\{jianyuan.guo, kai.han, yunhe.wang\}@huawei.com; c.xu@sydney.edu.au}
}

\maketitle

\begin{abstract}
Knowledge distillation is a widely used paradigm for inheriting information from a complicated teacher network to a compact student network and maintaining the strong performance. Different from image classification, object detectors are much more sophisticated with multiple loss functions in which features that semantic information rely on are tangled. In this paper, we point out that the information of features derived from regions excluding objects are also essential for distilling the student detector, which is usually ignored in existing approaches. In addition, we elucidate that features from different regions should be assigned with different importance during distillation. To this end, we present a novel distillation algorithm via decoupled features~(DeFeat) for learning a better student detector. Specifically, two levels of decoupled features will be processed for embedding useful information into the student, \ie, decoupled features from neck and decoupled proposals from classification head. Extensive experiments on various detectors with different backbones show that the proposed DeFeat is able to surpass the state-of-the-art distillation methods for object detection. For example, DeFeat improves ResNet50 based Faster R-CNN from 37.4\% to 40.9\% mAP, and improves ResNet50 based RetinaNet from 36.5\% to 39.7\% mAP on COCO benchmark. Our implementation is available at \href{https://github.com/ggjy/DeFeat.pytorch}{https://github.com/ggjy/DeFeat.pytorch}.
\end{abstract}

\section{Introduction}
As one of the fundamental computer vision tasks, object detection has attracted increasing attention in various real-world applications including autonomous driving and surveillance video analysis. Recent advances of deep learning introduce many convolutional neural network based solutions to object detection. The backbone of a detector is often composed of heavy convolution operations to produce intensive features that is critical to the detection accuracy. But doing so inevitably results in a sharp increase in the cost of computing resource and an apparent decrease in detection speed. Techniques such as quantization~\cite{han2015deep,wu2016quantized,fullyquan2019cvpr,quanmimic2018eccv,yang2020searching}, pruning~\cite{alvarez2016learning,guo2016dynamic,quan2015nips}, network design~\cite{pelee,efficientdet,hitdet,han2020ghostnet} and knowledge distillation \cite{distillfinegrained,efficientkd2017nips} have been developed to overcome this dilemma and achieve an efficient inference on detection task. 
We are particularly interested in knowledge distillation \cite{hinton2015distill}, as it provides an elegant way to learn a compact student network when a performance proven teacher network is available. 
Classical knowledge distillation methods are firstly developed for the classification task to decide \emph{which} category the image belongs to. The information from soft label outputs \cite{hinton2015distill,kim2018paraphrasing,TAKD,fang2019data} or intermediate features \cite{ahn2019variational,heo2019knowledge,zagoruyko2016paying} of a well-optimized teacher network have been well exploited to learn the student networks, but these methods cannot be directly extended to the detection task which needs to further figure out \emph{where} the objects are.

\begin{figure}[t]
	\centering
	\includegraphics[width=\columnwidth]{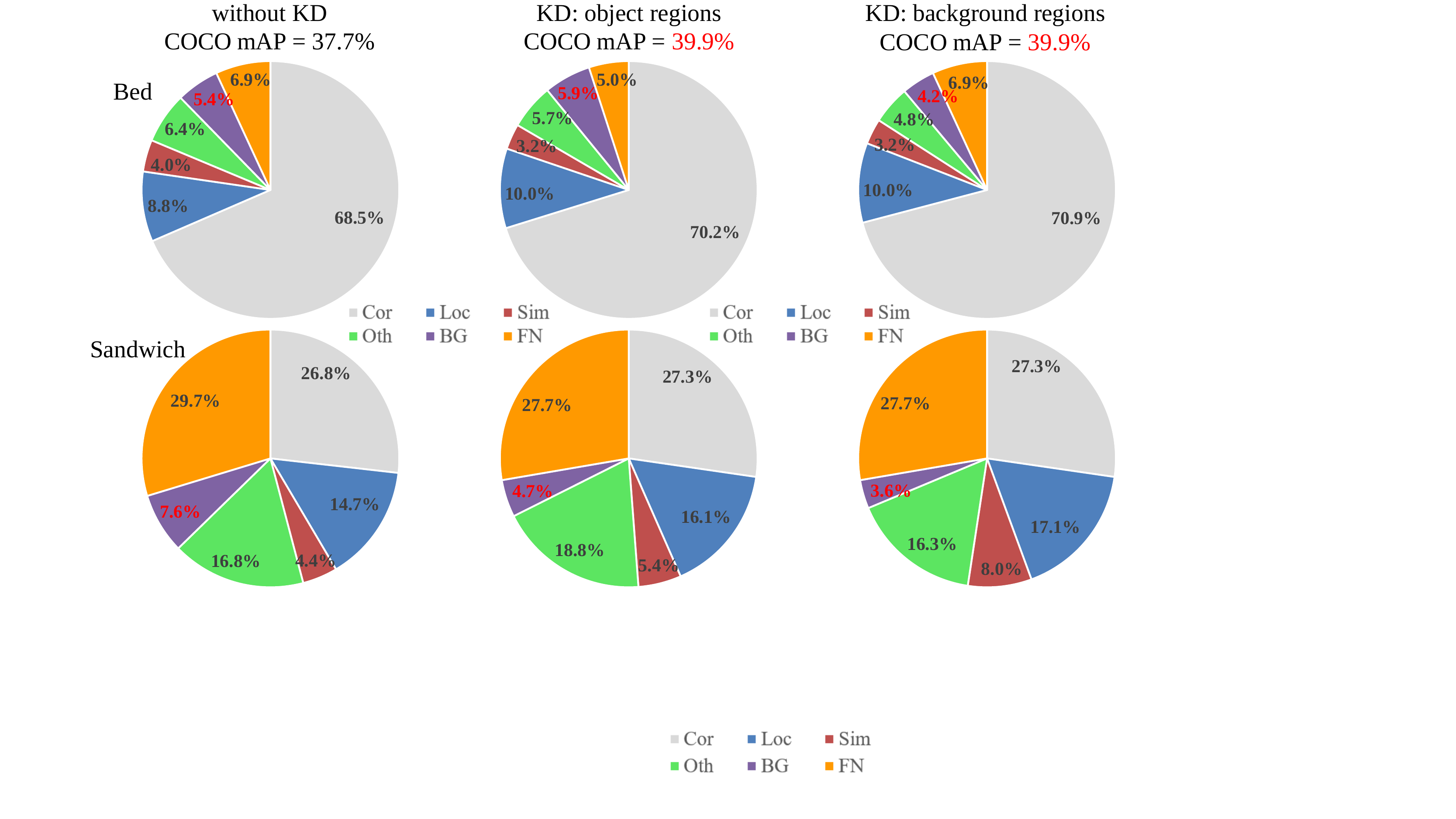}
	\caption{\small{Error analyses of different distillation methods on COCO minival. KD via background regions alleviates the false positive rate and achieves comparable result with KD via object regions. \textbf{Cor}: correct class (IoU $>$ 0.5). \textbf{Loc}: correct class but misaligned box (0.1 $<$ IoU $<$ 0.5). \textbf{Sim}: wrong class but correct supercategory (IoU $>$ 0.1). \textbf{Oth}: wrong class (IoU $>$ 0.1). \textbf{BG}: background false positives (IoU $<$ 0.1). \textbf{FN}: false negatives (remaining errors).}}
	\label{fig:intro_pie}
\end{figure}

\begin{figure*}[t]
	\centering
	\includegraphics[width=\linewidth]{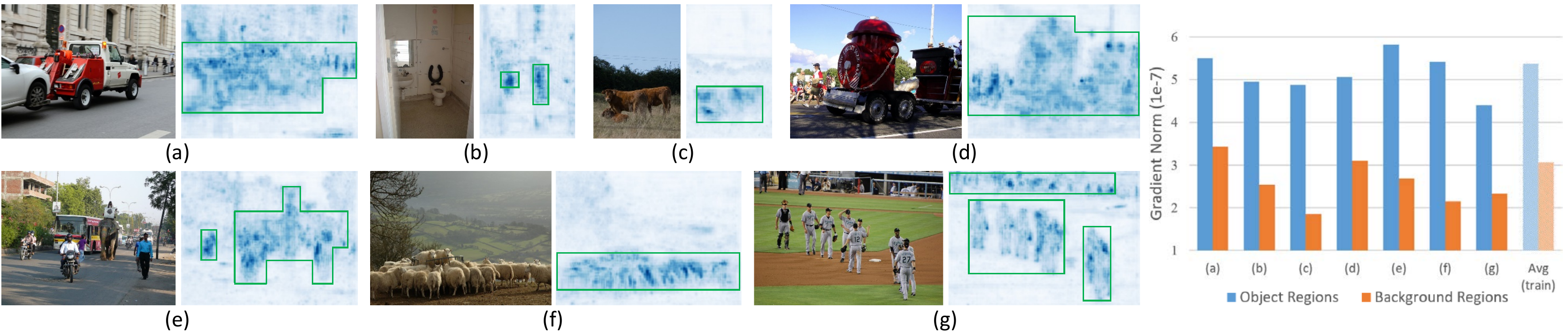}
	\vspace{-0.5cm}
	\caption{\small{Left: $L_2$-Norm of the gradient in intermediate neck feature during back propagation, the darkest blue indicates the largest norm value. Images are randomly selected from COCO training set, and object regions are marked with the green box. Right: Average $L_2$-Norm of object and background regions. Avg (train) indicates the average norm of all images from COCO training set.}}
	\label{fig:vis_gradient}
\end{figure*}

There are a few attempts investigating knowledge distillation in the object detection task. For example, FGFI \cite{distillfinegrained} asks the student network to imitate the teacher network on the near object anchor locations. TADF \cite{sunruoyu} distills the student via Gaussian masked object regions in neck features and positive samples in detection head.
These works only distilled knowledge from object regions, as background regions were supposed to be not of interests in the detection task. Intuitively, during the distillation, background regions might introduce a large amount of noise and they have rarely been explored. But there lacks a thorough analysis of background regions when conducting the distillation. The hasty decision of throwing away background regions thus might not be wise. Most importantly, background information has already been proven to be helpful for visual recognition \cite{torralba2003contextual,sarwar2018detecting,de2019does,guo2019beyond}.
Instead of guessing that background regions are useless or even harmful for distillation, it is time to have a fair and thorough analysis of the background and let the facts speak for themselves.

We first examine the roles of object and background regions in knowledge distillation by comparing two approaches: (i) distilling only via object-region FPN features and (ii) distilling only via background-region FPN features.
It was taken for granted that the student would not be enhanced significantly when distilled via the background regions from teacher detector, since the background is less informative and noisy \cite{distillfinegrained}. 
However, after extensive experiments on various models and datasets, we observe a surprising result that distilling student only via background-region features can also enhance the student remarkably and even, achieve comparable results with that of distillation via object regions (Figure~\ref{fig:coefficient}). We further explore where the performance improvement comes from by distilling background features. Taking two classes from COCO as an example (see Figure~\ref{fig:intro_pie}), we conduct the error analysis \cite{cocoerror} and find that distillation via background regions effectively reduces the number of background false positives.

The above evidence points to the conclusion that background regions can actually be a complementary to that distillation on object regions. Except that, prior literature has shown that there is a strong relationship between objects and background~\cite{torralba2003contextual,zhu2016object}. The object likelihood~\cite{torralba2003contextual} can be written as $P(O|V_o,V_b)$=$P(O|V_b)\frac{P(V_o|O,V_b)}{P(V_o|V_b)}$ ($V_o$ and $V_b$ are features of object region and background, respectively). All probabilities are related to background information which provides an estimate of the likelihood of finding an object (for example, one is unlikely to find a car in the room). The background-based priors vary from different images~\cite{zhu2016object}, thus we need to learn background feature for better prediction. However, the promising expectation above was failed to be justified by previous works~\cite{efficientkd2017nips,li2017mimicking} that take both object and background regions into account. Although they leveraged both types of regions, the student was not significantly improved compared to those only using object regions, which seems to agree with the phenomenon indicated by \cite{distillfinegrained}. Either the object or background regions can independently benefit the object detection through the distillation, but once they are integrated together, the performance drops unexpectedly. The reason could be that their methods integrate these two types of regions directly. From the gradient point of view, we illustrate the discordance between object and background regions in Figure~\ref{fig:vis_gradient}. Images in the left column are randomly selected from COCO training set, and images in the right column are their corresponding gradients of neck features in student detector. We can observe that the magnitude of gradients from object regions are consistently larger than that from background regions. This therefore reminds us of different importance of object regions and background regions during the distillation. 

Based on these insightful observations, we propose to decouple the features used for knowledge distillation and highlight their unique importance during the distillation. Two levels of features are included, \ie, FPN features and RoI-aligned features. The FPN features are split into object and background parts using the ground-truth mask, and the mean square error loss is applied between teacher and student. The RoI-aligned features are also decoupled into positive and negative parts using teacher's predicted region proposals. The classification logits generated based on these decoupled RoI-aligned features are distilled using the KL divergence loss. The resulting DeFeat algorithm can be adaptively incorporated into both one-stage and two-stage detectors to improve the detection accuracy. To validate our method, we conduct extensive experiments on Faster R-CNN \cite{faster-rcnn} and RetinaNet \cite{focal} under various scenarios including distillation on shallow student and narrow student on two common detection benchmarks PASCAL VOC \cite{voc} and COCO \cite{coco}. In particular, our DeFeat improves ResNet50 based FPN from 37.4\% to 40.9\% mAP, and ResNet50 based RetinaNet from 36.5\% to 39.7\% mAP on COCO benchmark.

\section{Related Work}
\noindent \textbf{Object detection} is considered as one of the most challenging vision tasks which aims at finding out \emph{what} and \emph{where} the objects are when given an image. In the past few years, noticeable improvements in accuracy have been made in both one-stage~\cite{yolov1,ssd,focal,cornernet,centernet,refinenet} and two-stage~\cite{faster-rcnn,mask-rcnn,fpn,relation,huang2017speed,cascade} detectors. 
Although detectors fitted with very deep backbone \cite{resnext,inceptionv4} have better detection accuracy, they are expensive in terms of computation cost and hard to deploy to mobile devices. 
There has been an interesting line of research that compresses large detection models by weight quantization \cite{fullyquan2019cvpr,quanmimic2018eccv}, representing the parameter weights with fewer bits. Pruning \cite{masana2016fly,ghosh2019deep,xie2020localization,tang2020scop,tang2021manifold} is another line of research that removes unimportant connections from a large pre-trained model to compress detector. Designing a detector coupled with lightweight backbone network \cite{pelee,thundernet,mobilenetv2,tinydsod,yang2020cars,you2020greedynas} is also a trend for faster detection speed. Besides, there is also a line of research that transfers knowledge from a large detector to a smaller detector \cite{efficientkd2017nips,distillfinegrained,li2017mimicking}, in which one can boost the performance of a small detector without designing new architectures.

\vspace{0.0cm}
\noindent \textbf{Knowledge distillation} (KD) has become one of the most effective techniques to compress large models into smaller and faster ones. KD was first proposed by Buciluǎ \etal~\cite{firstkd} and popularized by Hinton \etal~\cite{hinton2015distill} that transfers the dark knowledge from teacher network to student network through the soft outputs. FitNets \cite{fitnets} shows that activations \cite{heo2019knowledge} and features of intermediate layers \cite{passalis2018learning} can also be treated as knowledge to guide the student network. Since then, KD has been widely adopted in classification tasks~\cite{heo2019comprehensive,yim2017gift,anil2018large,similarity2019cvpr,chen2020learning,you2017learning,du2020agree}. Recently, there are several works which propose to compress object detector using knowledge distillation. Chen \etal~\cite{efficientkd2017nips} distills the student through all components (\ie, neck feature, classification head and regression head), but the imitation of entire feature maps and distillation in classification head both ignore the imbalance in foreground and background which could lead to a suboptimal result. Tang \etal \cite{sensetimebmvc} proposes adaptive distillation loss for one-stage detector to magnify loss on hard samples. 
Li \etal \cite{li2017mimicking} distills the features sampled from region proposals, however, only mimicking above regions could cause misguidance since the proposals can sometimes perform poorly. Wang \etal \cite{distillfinegrained} intends to distill the student with fine-grained features from foreground object regions. However, we find that the remaining background features are also critical for distilling a better student detector.

In summary, current distillation frameworks for object detection ignore the important roles of the background regions in intermediate features and negative region proposals in classification head. In this work, we identify that the object and background regions in FPN features are both practical for distillation and treating positive and negative proposals equally would withhold the detector of stronger performance. Therefore we first generate a binary mask to decouple the intermediate features and then distill the features accordingly. Meanwhile, we decouple the positive and negative proposals in classification head to further improve the generalization.

\section{Distillation via Decoupled Features}
Generally, an object detector consists of three or four components: (a) backbone for extracting semantic features; (b) neck for fusing multi-level features; (c) RPN for generating proposals (only in two-stage detectors); and (d) head for object classification and bounding box regression. The purpose of distillation is to imbue the student with dark knowledge inside the teacher, which can be features of intermediate layer or soft predictions of region proposals in classification head.
Define $\mathcal{S}$$\,\in\,$$\mathbb{R}^{H\times W\times C}$ and $\mathcal{T}$$\,\in\,$$\mathbb{R}^{H\times W\times C}$ as the intermediate features of student and teacher, respectively. The distillation via intermediate features can be formulated as:
\vspace{-0.1cm}
\begin{equation}
\mathcal{L}_{fea} =  \frac{\gamma}{2N}\sum_{h=1}^{H}{\sum_{w=1}^{W}{\sum_{c=1}^{C}{I({\phi({\mathcal{S}}_{h,w,c}})-{\mathcal{T}_{h,w,c}})^2}}},
\label{eq:one-neck}
\end{equation}
where $N=HWC$ is the total number of elements, $\gamma$ is used to control the scale of distillation loss, $\phi$ denotes the adaptation layer \cite{efficientkd2017nips} and $I$ denotes the imitation mask, \ie, Gaussian mask in \cite{sunruoyu} and fine-grained mask in \cite{distillfinegrained}. In previous works, only object regions are considered or the entire feature maps are distilled uniformly. 
Mask in methods that treat all regions uniformly \cite{efficientkd2017nips,li2017mimicking} can be seen as an all-one tensor.

Given $K$ region proposals output from RPN, the classification head needs to compute soft labels of all proposals. The distillation via soft predictions can be formulated as:
\begin{equation}
\mathcal{L}_{cls} = \frac{1}{K}\sum_{i=1}^{K}{\mathcal{L}_{CE}(y^s_i,Y_i)} + \frac{\lambda}{K}\sum_{i=1}^{K}{\mathcal{L}_{KL}(y^s_i,y^t_i)},
\label{eq:previous-cls}
\end{equation}
where the hyper-parameter $\lambda$ is used to balance different loss items, $\mathcal{L}_{CE}$ and $\mathcal{L}_{KL}$ denote the cross entropy loss and the KL divergence loss, respectively. $Y_i$ is the ground truth label of the $i$-th proposal, and the predictions of student and teacher detectors are $y^s_i$ and $y^t_i$, respectively.
The overall training targets of the student can be formulated as:
\begin{equation}
\mathcal{L} = \mathcal{L}_{fea} + \mathcal{L}_{cls} + \mathcal{L}_{reg} + \mathcal{L}_{rpn},
\label{eq:total-loss}
\end{equation}
where $\mathcal{L}_{reg}$ is the bounding box regression loss in detector head and $\mathcal{L}_{rpn}$ denotes the RPN loss in two-stage detector.

\begin{figure*}[t]
	\centering
	\includegraphics[width=\linewidth]{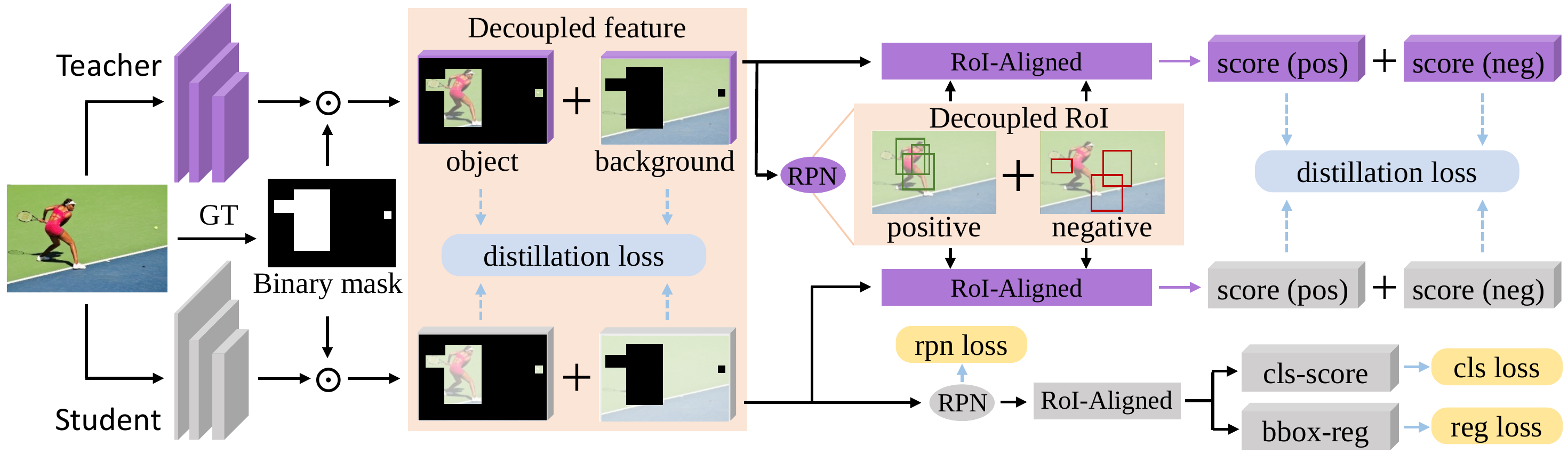}
	\caption{\small{Overview of the proposed distillation via decoupled features (DeFeat) framework. We decouple the regions in intermediate FPN features and the region proposals from RPN to distill the student detector. The terms ``score (pos)" and ``score (neg)" indicate the classification scores on positive and negative proposals, respectively.}}
	\label{fig:framework}
\end{figure*}

\subsection{Decouple Intermediate Features in Distillation}
Previous works either choose partial regions or use all regions but treat each location on intermediate features equally. In particular, FGFI \cite{distillfinegrained} presumed that background regions could introduce a large amount of noise and would impair the performance. However, this intuitive judgment is not consistent with what we have observed in experiments, as shown in Figure \ref{fig:intro_pie}.
Distillation via background only regions still achieves comparable results as distillation via object only regions.
We come to a conclusion that the background regions in intermediate features can complement the object regions and further help the training of student detector, but the remaining question is how to appropriately integrate these two types of regions in distillation. 

Based on the observations above, we propose to distill the student via decoupled features. Given the intermediate features of size $H\times W$, we first generate a binary mask $M$ according to the ground truth box $\rm{B}$:
\begin{equation}
M_{i,j} = \mathds{1}[(i,j)\in \rm{B}],
\end{equation}
where $M$$\,\in\,$$\{0,1\}^{H\times W}$, the value of location $(i,j)$ is 1 if it belongs to an object, and 0 otherwise. Specifically, if detectors contain the feature pyramid network (FPN) which can output multi-level features, we will assign each ground truth box to its corresponding level and generate the mask $M$ for each level accordingly. Then we use the generated binary mask to decouple the neck features, as shown in Figure \ref{fig:framework}. The intermediate feature distillation is formulated as:
\begin{equation}
\begin{aligned}
\mathcal{L}_{fea} = \frac{\alpha_{obj}}{2N_{obj}}\sum_{h=1}^{H}{\sum_{w=1}^{W}{\sum_{c=1}^{C}{M_{h,w}({\phi({\mathcal{S}}_{h,w,c}})-{\mathcal{T}_{h,w,c}})^2}}} \\
+ \; \frac{\alpha_{bg}}{2N_{bg}}\sum_{h=1}^{H}{\sum_{w=1}^{W}{\sum_{c=1}^{C}{(1-M_{h,w})({\phi({\mathcal{S}}_{h,w,c}})-{\mathcal{T}_{h,w,c}})^2}}},
\end{aligned}
\label{eq:kd-neck}
\end{equation}
where $N_{obj}$$\,=\,$$C\sum_{h=1}^H\sum_{w=1}^W{M_{w,h}}$ is the number of elements in object regions,  $N_{bg}$$\,=\,$$C\sum_{h=1}^H\sum_{w=1}^W{(1-M_{w,h})}$ is the number of elements in background regions. $\alpha_{obj}$ and $\alpha_{bg}$ are the loss coefficients for object and background regions, respectively. 
Through the ground-truth based mask, we decouple the features into object and background regions to distill both of them in a balanced manner.

\subsection{Decouple Region Proposals in Distillation}
Knowledge distillation via the soft predictions has been widely used in classification task, and can be useful for distilling the classification head in detection task. However, different from the classification task that there is no background category during training (\eg, CIFAR and ImageNet), the object and background categories in detection head can have extremely different numbers of proposals. We conduct experiments to explore the separate distillation losses of object (positive) proposals and background (negative) proposals as shown in Figure~\ref{fig:loss_proposal}. The distillation loss of positive proposals is consistently larger than that of negative proposals. If they are not properly balanced, the small gradients produced by background proposals can be drowned into the large gradients produced by positive ones, thus limiting further refinement. 
Besides, Table~\ref{table:proposal} shows that treating all proposals equally gets worse result compared to using negative only proposals.
Hence, we propose to decouple the region proposals into positive ones and negative ones towards the optimal convergence when distilling the classification head. We feed the region proposals produced by teacher detector into both teacher's and student's head to generate the category predictions $p^t$ and $p^s$ as shown in Figure \ref{fig:framework}. The positive proposals and negative proposals are processed separately in our method. Given the logits $z$ of positive proposals, we soften the predictions by a temperature $T_{obj}$ for teacher and student as following:
\begin{align}
p^{s,T_{obj}}(c \mid \theta^s)=\frac{exp(z_c^s/T_{obj})}{\sum_{j=1}^C{exp(z_j^s/T_{obj})}}, c\in Y
\label{eq:student probability} \\
p^{t,T_{obj}}(c \mid \theta^t)=\frac{exp(z_c^t/T_{obj})}{\sum_{j=1}^C{exp(z_j^t/T_{obj})}}, c\in Y
\label{eq:teacher probability}
\end{align}

\begin{table*}[t]
	\renewcommand\tabcolsep{5.0pt}
	\centering
	\caption{\small{Distillation results of both two-stage detector FPN and one-stage detector RetinaNet on COCO benchmark. ``ResNet152-R50-FPN" indicates that the teacher detector is ResNet152 based FPN, while the student detector is ResNet50 based FPN.}}
	\begin{tabular}{llcccc}
		\toprule[1pt]
		\specialrule{0em}{1pt}{0pt}
		Method & Distillation & mAP & $\rm{AP}_{S}$ & $\rm{AP}_{M}$ & $\rm {AP}_{L}$ \\ 
    		\hline
		Teacher & ResNet152-FPN & 41.3 & 24.4 & 45.3 & 54.0 \\ 
		Student & ResNet50-FPN & 37.4 & 21.8 & 41.0 & 47.8 \\
		\hline
		KD: all-neck & ResNet152-R50-FPN & 40.1 ($\blacktriangle$2.7) & 22.8 & 44.0 & 53.3 \\
		KD: decoupled-neck & ResNet152-R50-FPN & 40.4 ($\blacktriangle$3.0) & 23.4 & 44.4 & 53.1 \\
		KD: decoupled-neck + all-cls & ResNet152-R50-FPN & 40.5 ($\blacktriangle$3.1) & 23.6 & 44.6 & 53.1 \\
		KD: decoupled-neck + decoupled-cls & ResNet152-R50-FPN & 40.8 ($\blacktriangle$3.4) & 23.5 & 44.8 & 53.3 \\
		KD: backbone + decoupled-neck + decoupled-cls & ResNet152-R50-FPN & \textbf{40.9} ($\blacktriangle$3.5) & \textbf{23.6} & \textbf{44.8} & \textbf{53.5} \\ 
		\specialrule{0em}{-0.5pt}{0pt}
		\midrule[1pt]
		\specialrule{0em}{1pt}{1pt}
		Teacher & ResNet152-RetinaNet & 40.5 & 24.1 & 44.7 & 53.4 \\ 
		Student & ResNet50-RetinaNet & 36.5 & 20.9 & 40.2 & 47.0 \\
		\hline
		KD: all-neck & ResNet152-R50-RetinaNet & 39.1 ($\blacktriangle$2.6) & 22.1 & 43.1 & 52.3 \\
		KD: decoupled-neck & ResNet152-R50-RetinaNet & 39.5 ($\blacktriangle$3.0) & 23.3 & 43.4 & 52.7 \\
		KD: backbone + decoupled-neck & ResNet152-R50-RetinaNet & \textbf{39.7} ($\blacktriangle$3.2) & \textbf{23.4} & \textbf{43.6} & \textbf{52.9} \\
		\specialrule{0em}{-0.5pt}{0pt}
		\bottomrule[1pt]
	\end{tabular}
	\label{table:main}
\end{table*}

where $\theta^s$ and $\theta^t$ denote the parameters of the student and the teacher, respectively. $Y=\{1,2,...,C\}$ are the classes of detection benchmark. For proposals belonging to background regions, we soften the predictions by a temperature $T_{bg}$ for teacher and student similar to equations above. To distill the student with knowledge from teacher detectors, we use the Kullback Leibler (KL) divergence written as:
\vspace{-0.1cm}
\begin{equation}
\begin{aligned}
\mathcal{L}_{cls} = \frac{\beta_{obj}}{K_{obj}}\sum_{i=1}^K{b_i\mathcal{L}_{KL}(p^{s,T_{obj}}_i,p^{t,T_{obj}}_i)} \\  
+ \; \frac{\beta_{bg}}{K_{bg}}\sum_{i=1}^K{(1-b_i)\mathcal{L}_{KL}(p^{s,T_{bg}}_i,p^{t,T_{bg}}_i)}
\end{aligned}
\label{eq:kd-cls}
\end{equation}
\begin{equation}
\mathcal{L}_{KL}(p^{s,T},p^{t,T})=T^2\sum_{c=1}^C{p^{t,T}(c\,|\,\theta^t)\log{\frac{p^{t,T}(c\,|\,\theta^t)}{p^{s,T}(c\,|\,\theta^s)}}}
\end{equation}
where $b_i$$\,\in\,$$\{0,1\}$ is the binary label of $i$-th proposal with respect to ground truth object. $\beta_{obj}$ and $\beta_{bg}$ are the coefficients of positive and negative samples, respectively. $K_{obj}$$\,=\,$$\sum_i{b_i}$ and $K_{bg}$$\,=\,$$\sum_i{(1-b_i)}$ are the numbers of positive and negative proposals, respectively. And we multiply the distillation loss by $T^2$ to ensure the scale of gradient magnitudes.

\section{Experiments}
\subsection{Datasets and Metrics}
\noindent \textbf{COCO} \cite{coco} is a challenging benchmark in object detection which contains 80 object classes. Our training set is the union of 80k training images and 35k subset of validation images (trainval35k), and the validation set is the remaining 5k validation images (minival). We consider Average Precision as evaluation metric, \ie, $\rm mAP$, $\rm {AP}_{50}$, $\rm {AP}_{75}$, $\rm {AP}_{S}$, $\rm {AP}_{M}$ and $\rm {AP}_{L}$. The last three measure performance with respect to objects with different scales.

\begin{table}[t]
	\renewcommand\arraystretch{1.0}
	\small
	\centering
	\caption{\small{Comparison with state-of-the-art methods on COCO.}}
	\setlength{\tabcolsep}{4.0pt}{
		\begin{tabular}{lccccc}
			\toprule[1pt]
			\specialrule{0em}{1pt}{0pt}
			Method & Distillation & mAP & $\rm{AP}_{S}$ & $\rm{AP}_{M}$ & $\rm {AP}_{L}$ \\ 
			\hline
			Teacher & R152-FPN & 41.3 & 24.4 & 45.3 & 54.0 \\
			Student & R50-FPN & 37.4 & 21.8 & 41.0 & 47.8 \\
			FGFI & R152-R50-FPN & 39.9 & 22.9 & 43.6 & 52.8 \\
			TADF & R152-R50-FPN & 40.1 & 23.0 & 43.6 & 53.0 \\
			DeFeat & R152-R50-FPN & \textbf{40.9} & \textbf{23.6} & \textbf{44.8} & \textbf{53.5} \\
			\hline
			Teacher & R50-FPN & 37.4 & 21.8 & 41.0 & 47.8 \\
			Student & R50(1/4)-FPN & 29.1 & 16.2 & 31.1 & 38.5 \\
			FGFI & R50-R50(1/4)-FPN & 31.8 & 17.1 & 34.2 & 43.0  \\
			DeFeat & R50-R50(1/4)-FPN & \textbf{33.0} & \textbf{18.2} & \textbf{35.5} & \textbf{44.0} \\
			\hline
			Teacher & R152-RetinaNet & 40.5 & 24.1 & 44.7 & 53.4 \\
			Student & R50-RetinaNet & 36.5 & 20.9 & 40.2 & 47.0 \\
			FGFI & R152-R50-RetinaNet & 38.9 & 21.9 & 42.5 & 52.2 \\
			DeFeat & R152-R50-RetinaNet & \textbf{39.7} & \textbf{23.4} & \textbf{43.6} & \textbf{52.9} \\
			\specialrule{0em}{-0.5pt}{0pt}
			\bottomrule[1pt]
		\end{tabular}
	}
	\vspace{-0.2cm}
	\label{table:coco-sota}
\end{table}

\begin{table}[t]
	\renewcommand\arraystretch{1.0}
	\small
	\centering
	\caption{\small{Comparison with state-of-the-art methods on VOC.}}
	\setlength{\tabcolsep}{6pt}{
		\begin{tabular}{lccccc}
			\toprule[1pt]
			\specialrule{0em}{1pt}{0pt}
			Method & Distillation & mAP \\ 
			\hline
			Teacher & R152-FPN & 82.69 \\
			Student & R50-FPN & 80.53 \\
			FGFI \cite{distillfinegrained} & R152-R50-FPN & 81.57 \\
			TADF \cite{sunruoyu} & R152-R50-FPN & 81.71 \\
			DeFeat & R152-R50-FPN & \textbf{82.28} \\
			\hline
			Teacher & R101-FPN & 82.13 \\
			Student & R50-FPN & 80.53 \\
			FGFI \cite{distillfinegrained} & R101-R50-FPN & 81.02 \\
			FGFI + PAD \cite{prime} & R101-R50-FPN & 81.25 \\
			Mimic \cite{li2017mimicking} & R101-R50-FPN & 80.90  \\
			Mimic + PAD \cite{prime} & R101-R50-FPN & 81.11  \\
			DeFeat & R101-R50-FPN & \textbf{81.47} \\
			\specialrule{0em}{0pt}{-0.5pt}
			\bottomrule[1pt]
		\end{tabular}
	}
	\vspace{-0.4cm}
	\label{table:voc-sota}
\end{table}

\noindent \textbf{Pascal VOC} \cite{voc} contains 20 object classes. Our training set is the union of VOC 2007 trainval (5K) and VOC 2012 trainval (11K), and the validation set is the VOC 2007 test (4.9K). We report the mAP scores using IoU at 0.5.

\subsection{Implementation Details}
All experiments are performed on 8 Tesla V100 GPUs. Our implementation is based on mmdetection \cite{mmdetection} with Pytorch framework \cite{pytorch}. We use SGD optimizer with a batch size of 4 images per GPU, all models are trained for 12 epochs, known as 1$\times$ schedule. The input image is resized such that its shorter side has 800 pixels on COCO and 600 pixels on VOC. The initial learning rate is set as 0.04 and 0.02 for FPN \cite{fpn} and RetinaNet \cite{focal}, respectively. And the learning rate is divided by 10 at the 8-th and 11-th epochs. We set momentum as 0.9 and weight decay as 0.0001.

\subsection{Main Results}

We first verify the effectiveness of our proposed DeFeat on typical two-stage detection framework FPN \cite{fpn} on COCO \cite{coco} benchmark, as shown in Table \ref{table:main}. ResNet50 based FPN is chosen as the student detector and ResNet152 based FPN as the teacher detector. ``All-neck" indicates that the student is distilled via treating all regions in FPN features equally as illustrated in Equation \ref{eq:one-neck}. ``Decoupled-neck" means the student is distilled via decoupled FPN features as illustrated in Equation \ref{eq:kd-neck}. ``All-cls" indicates that all region proposals are treated equally in distillation. ``Decoupled-cls" denotes that region proposals are decoupled into positive (object) and negative (background) as illustrated in Equation~\ref{eq:kd-cls}. ``Backbone" means the backbone features are also distilled. Directly distilling all regions in FPN features achieves 40.1\% mAP, and the decoupled FPN features can further improve student detector by 0.3\% mAP. Student distilled via decoupled FPN features and RPN proposals achieves a higher result, and our decoupled proposals can boost the result from 40.5\% to 40.8\% mAP. Further adopting the backbone features in distillation will achieve 40.9\% mAP on COCO benchmark, bringing 3.5\% gains compared to the student baseline model. In addition, we also conduct the experiments on typical one-stage detection framework RetinaNet \cite{focal}, our ResNet152-R50-RetinaNet improves the baseline counterpart from 36.5\% to 39.7\% mAP on COCO. These results clearly elucidate the versatility and generality of our proposed DeFeat in both one-stage and two-stage detectors.

\subsection{Comparison with State-of-the-art Methods}

\begin{figure}[t]
	\centering
	\includegraphics[width=\columnwidth]{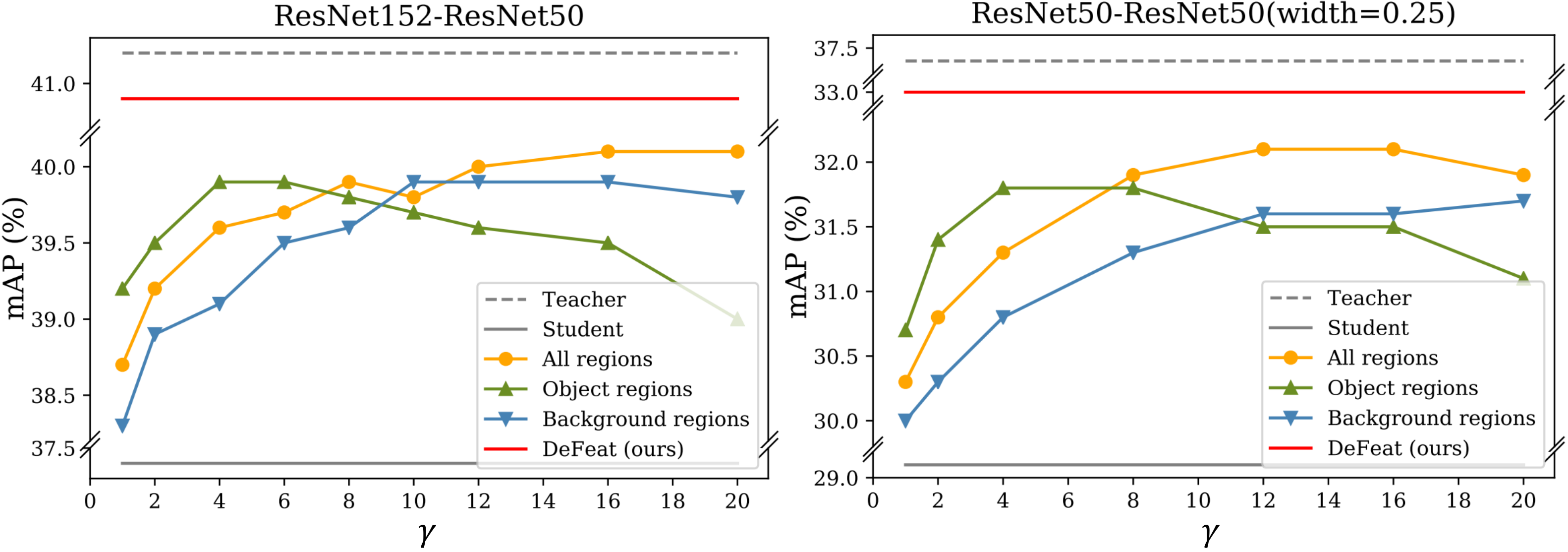}
	\caption{\small{Comparisons of different distillation regions and various distillation loss coefficients on COCO. Left: ResNet152 based FPN teacher distills a shallower ResNet50-FPN student. Right: ResNet50-FPN teacher distills a narrower Quartered-ResNet50-FPN student (number of backbone channels is quartered).}}
	\vspace{-0.2cm}
	\label{fig:coefficient}
\end{figure}

\begin{figure}[h!]
	\centering
	\includegraphics[width=\linewidth]{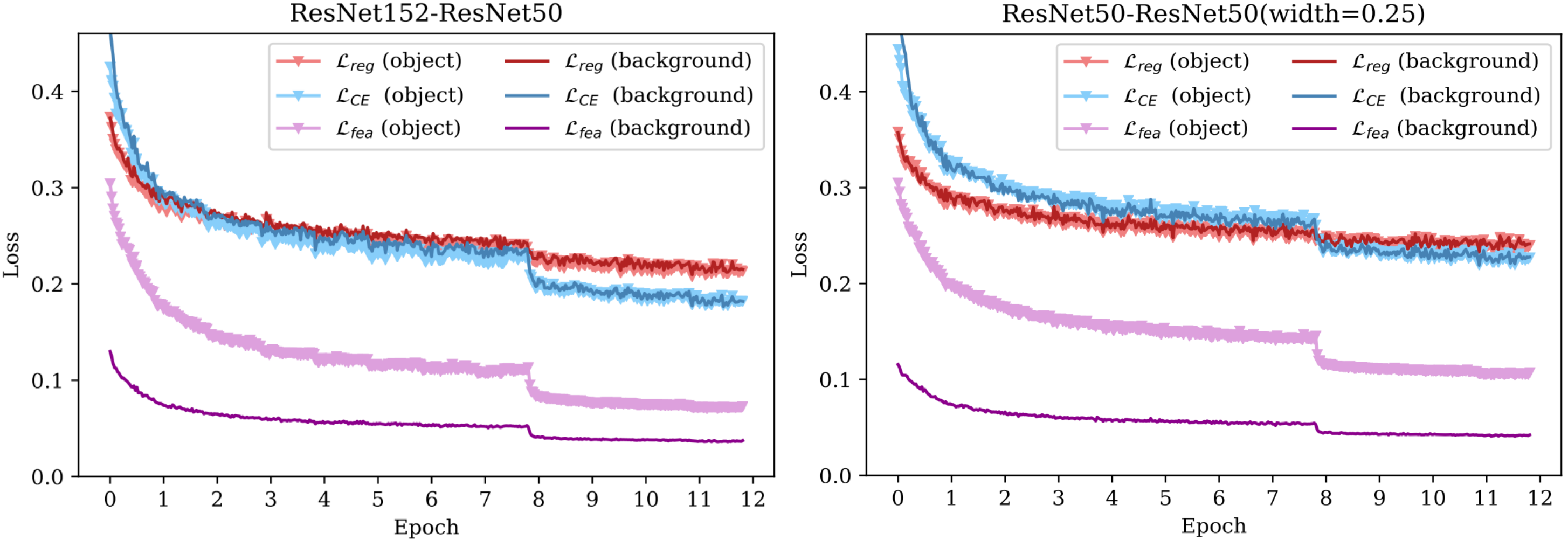}
	\caption{Training loss of distillation via neck features on COCO. Legend ``object" denotes using object only regions and ``background" denotes using background only regions.}
	\vspace{-0.2cm}
	\label{fig:loss_neck}
\end{figure}

\begin{figure}[h!]
	\centering
	\includegraphics[width=\linewidth]{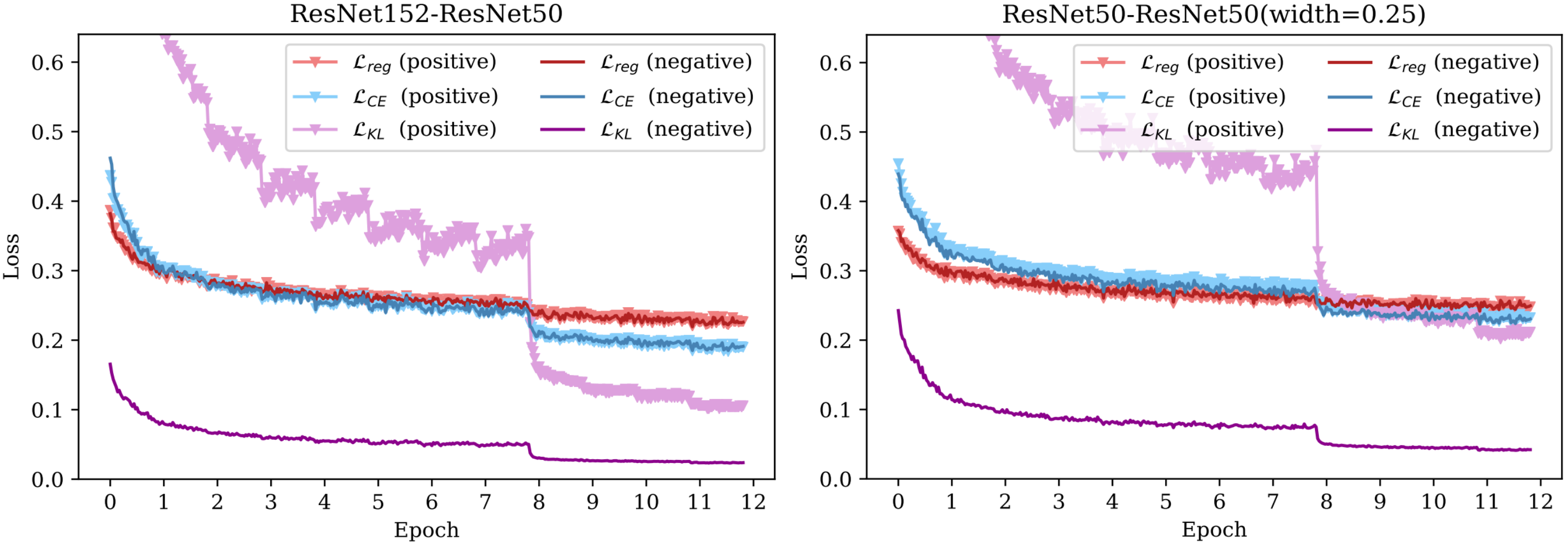}
	\caption{Training loss of distillation via region proposals on COCO. Legend ``positive" denotes using positive only proposals and ``negative" denotes using negative only proposals.}
	\vspace{-0.3cm}
	\label{fig:loss_proposal}
\end{figure}

Comparison of the results obtained with other state-of-the-art distillation methods on COCO \cite{coco} benchmark and Pascal VOC \cite{voc} benchmark are shown in Table \ref{table:coco-sota} and Table \ref{table:voc-sota}, respectively. Mimic \cite{li2017mimicking} uses all regions in neck features to distill the student detector. FGFI \cite{distillfinegrained} distills the student detector via partial fine-grained regions in neck features. TADF \cite{sunruoyu} distills the student detector via Gaussian masked regions in neck features and positive only region proposals in both classification and regression head. PAD \cite{prime} proposes to estimate the weight of each region proposal during distillation. For fair comparison, we re-implement their methods, and most of them are slightly higher than the results in original paper. Our proposed DeFeat can be easily applied to the two most mainstream frameworks and consistently improve the performances of student detectors under various circumstances, \eg, different backbones, shallower student and narrower student. FGFI achieves 39.9\% mAP and TADF obtains 40.1\% mAP on COCO benchmark. However, these two methods both ignore the important roles of background regions in neck features. Our distillation via decoupled features outperforms the FGFI by 1.0\% mAP, and surpasses the TADF by 0.8\% mAP, which indicates the effectiveness of the proposed method. Furthermore, our ResNet152-R50-FPN boosts the baseline model from 80.5\% to 82.3\% mAP, and ResNet101-R50-FPN improves the baseline model from 80.5\% to 81.5\% mAP on Pascal VOC benchmark, which outperforms other distillation methods apparently. To be specific, PAD~\cite{prime} uses a stronger baseline implemented on detectron2 (ms-train, 17 epoch, and 1200$\times$800 size). We report the results on mmdetection (ss-train, 12 epoch, and 1000$\times$600 size) for fair comparisons with FGFI and TADF here.

\subsection{Ablation Study}
\vspace{0.cm}
\noindent \textbf{Impact of background regions in neck features.} We investigate the object and background regions in neck features under two circumstances: (1) ResNet152 based FPN as teacher detector and ResNet50 based FPN as student detector; (2) ResNet50 based FPN as teacher detector and Quartered-ResNet50 based FPN (number of backbone channel is quartered, 64.28\% Top-1 accuracy on ImageNet) as student detector. Figure \ref{fig:coefficient} shows the corresponding results on COCO minival. We combine distillation loss based on FG-Mask \cite{distillfinegrained} with original detection loss to guide the learning of student detector. The orange line indicates that the student detector is distilled via all regions in teacher's neck features. The green line indicates that the student is distilled via object only regions. The blue line denotes that the student is distilled via background only regions. $\gamma$ is the coefficient in Equation~\ref{eq:one-neck}. 
When the coefficient of distillation loss is small, learning from the object only (fine-grained) regions achieves better results. While the coefficient increases, performance improvement brought by background (non fine-grained) regions increases and then surpasses the result of mimicking object only regions. And the best results of mimicking object only regions and mimicking background only regions are comparable. Thus, we can come to the conclusion that both object regions and background regions in neck features are critical and have detrimental effects on the distillation of student detector.
And we set $\alpha_{obj}$=4 and $\alpha_{bg}$=16 in Equation \ref{eq:kd-neck} accordingly.

\begin{table}[t]
	\renewcommand\tabcolsep{5.0pt}
	\centering
	\caption{\small{Comparison of various region selection masks on COCO.}}
	\begin{tabular}{lccc}
		\toprule[1pt]
		\specialrule{0em}{1pt}{0pt}
		Model & Region & Mask & mAP \\ \hline
		R152-R50-FPN & obj & FG-Mask & 39.9 \\
		R152-R50-FPN & obj & Gaussian-Mask & 39.8 \\
		R152-R50-FPN & obj & GT-Mask & 39.9 \\
		R152-R50-FPN & obj + bg & FG-Mask & 40.4 \\
		R152-R50-FPN & obj + bg & GT-Mask & 40.4 \\
		R152-R50-FPN & obj + bg & Random-Mask & 40.0 \\
		R152-R50-FPN & obj + bg & w/o Mask & 40.1 \\
		\hline
		R101-R50-FPN & obj & FG-Mask & 38.9 \\
		R101-R50-FPN & obj & Gaussian-Mask & 38.7 \\
		R101-R50-FPN & obj & GT-Mask & 38.8 \\
		\hline
		R50-R50(1/4)-FPN & obj & FG-Mask & 31.8 \\
		R50-R50(1/4)-FPN & obj & Gaussian-Mask & 31.5 \\
		R50-R50(1/4)-FPN & obj & GT-Mask & 31.7 \\
		\specialrule{0em}{-0.5pt}{0pt}
		\bottomrule[1pt]
	\end{tabular}
	\vspace{-0.2cm}
	\label{table:mask}
\end{table}

\begin{table}[t]
	\renewcommand\arraystretch{1.0}
	\centering
	\caption{\small{Ablation study on the effects of positive and negative region proposals for R152-R50-FPN on COCO.}}
	\setlength{\tabcolsep}{3.6pt}{
		\begin{tabular}{lccc|lccc}
			\toprule[1pt]
			\specialrule{0em}{1pt}{0pt}
			\multicolumn{3}{l}{Teacher (R152-FPN)} & 41.3 & Proposal & $\beta_{bg}$ & $T_{bg}$ & mAP \\
			\cline{5-8}
			\multicolumn{3}{l}{Student (R50-FPN)} & 37.4 & negative & 4 & 1 & 38.3 \\
			\cline{1-4}
			Proposal & $\beta_{obj}$ & $T_{obj}$ & mAP & negative & 2 & 1 & 38.6 \\
			\cline{1-4}
			positive & 1 & 1 & 35.2 & negative & 1 & 1 & 38.4  \\
			positive & 0.1 & 1 & 37.7 & negative & 1 & 2 & 38.2  \\
			positive & 0.1 & 3 & 37.9 & sub-neg. & 1 & 1 & 38.2  \\
			positive & 0.05 & 3 & 38.1 & negative & 0.1 & 1 & 37.4  \\
			\hline
		\end{tabular}}
	\setlength{\tabcolsep}{4.25pt}{
		\hspace*{-1.06pt}
		\begin{tabular}{lcccccc}
			Proposal & $\beta_{obj}$ & $\beta_{bg}$ & $T_{obj}$ & $T_{bg}$ & $\lambda$ & mAP \\ \hline
			positive + sub-neg. & - & - & - & - & 1 & 37.4 \\
			positive + negative & - & - & - & -  & 1 & 38.2 \\
			positive + negative & - & - & - & -  & 0.1 & 38.1 \\
			positive + negative & 0.05 & 2 & 1 & 1 & - & 38.6 \\
			positive + negative & 0.05 & 2 & 3 & 1 & - & 38.9 \\
			\specialrule{0em}{-0.5pt}{0pt}
			\bottomrule[1pt]
		\end{tabular}
	}
	\vspace{-0.2cm}
	\label{table:proposal}
\end{table}

\begin{table}[t]
	\renewcommand\arraystretch{1.0}
	\centering
	\caption{\small{Ablation study on shared proposals on COCO.}}
	\begin{tabular}{lcc}
		\toprule[1pt]
		\specialrule{0em}{1pt}{0pt}
		Model & Proposal & mAP \\ \hline
		R152-R50-FPN, decoupled-cls & Teacher & 38.9 \\
		R152-R50-FPN, decoupled-cls & Student & 38.7 \\
		R50-R50(1/4)-FPN, decoupled-cls & Teacher & 31.5 \\
		R50-R50(1/4)-FPN, decoupled-cls & Student & 31.2 \\
		\specialrule{0em}{-0.5pt}{0pt}
		\bottomrule[1pt]
	\end{tabular}
	\vspace{-0.2cm}
	\label{table:shared}
\end{table}

In addition, we find that the distillation loss of background regions is much smaller than that of object regions during training, which indicates that background regions have smaller gradients compared to object regions. Figure \ref{fig:loss_neck} shows the training losses (\ie, regression loss, classification loss and distillation loss) of mimicking object only and background only regions. Besides, the values of classification and regression losses are about four times larger than that of distillation via object regions, which should be the reason for why ``Object regions" gets the best performance at $\gamma=4$ in Figure \ref{fig:coefficient}. This can also explain that background regions need larger loss weight in training phase.

\begin{figure*}[t]
	\centering
	\includegraphics[width=\linewidth]{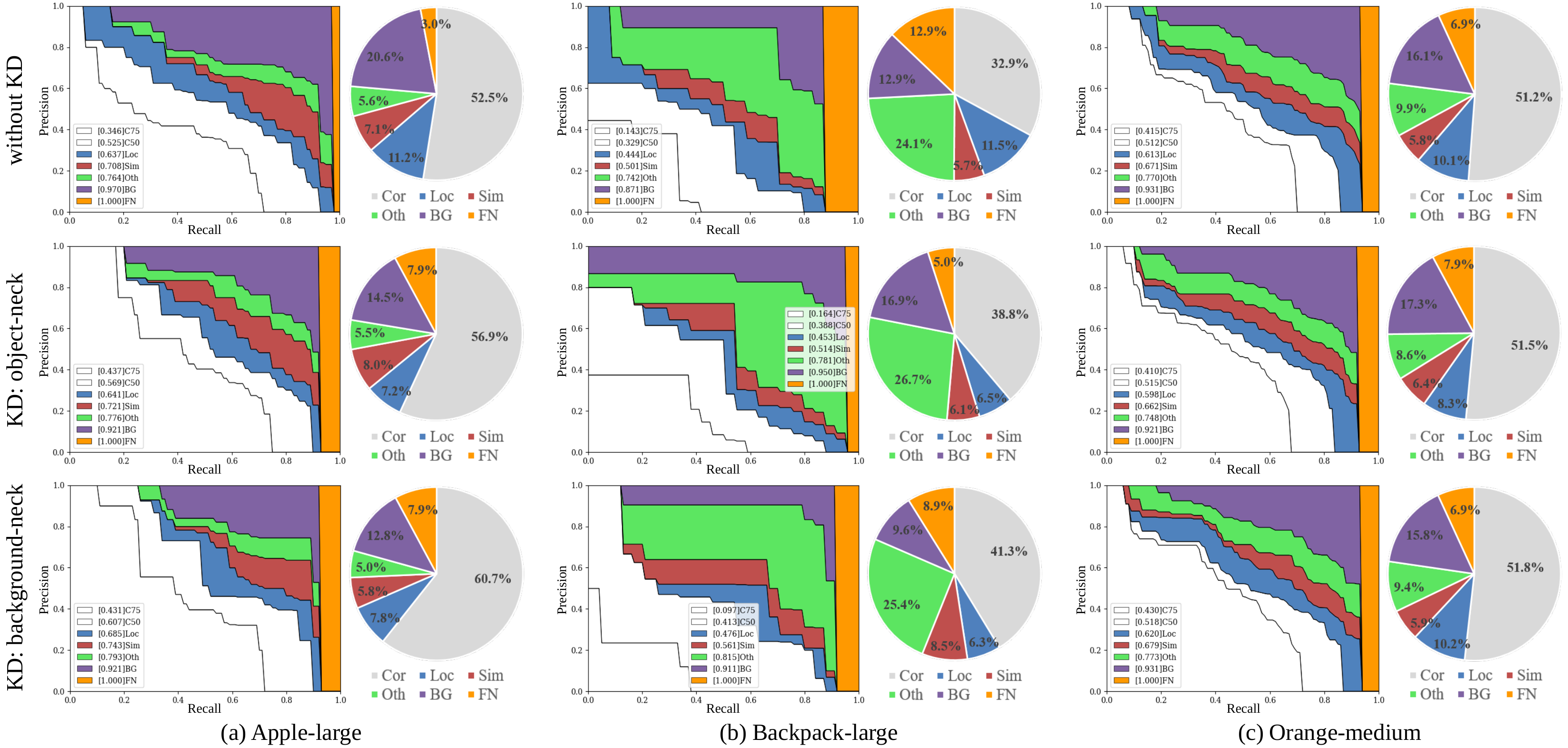}
	\vspace{-0.5cm}
	\caption{\small{Precision-Recall curves and error analyses of different distillation methods on COCO benchmark. For each case, top figures correspond to raw student model, middle figures correspond to distillation via object only regions, bottom figures correspond to distillation via background only regions.}}
	\vspace{-0.2cm}
	\label{fig:pie_exp}
\end{figure*}

\vspace{0.cm}
\noindent \textbf{Evaluation of different region selection masks.}
Here we evaluate several region selection masks, namely FG-Mask from \cite{distillfinegrained}, Gaussian-Mask from \cite{sunruoyu}, GT-Mask that directly leverages the ground truth boxes and Random-Mask which is generated randomly. Table \ref{table:mask} depicts the corresponding results. We can find that simply choosing scaled ground truth boxes as imitation regions achieves similar result compared to using fine-grained regions. And the distillation via object regions selected by Gaussian-Mask \cite{sunruoyu} obtains worse result. R152-R50-FPN achieves 40.1\% mAP by treating all regions equally, decoupled regions can further boost the result by 0.3\% mAP, while Random-Mask leads to a slight decrease of performance.

\begin{figure*}[t]
	\centering
	\includegraphics[width=\linewidth]{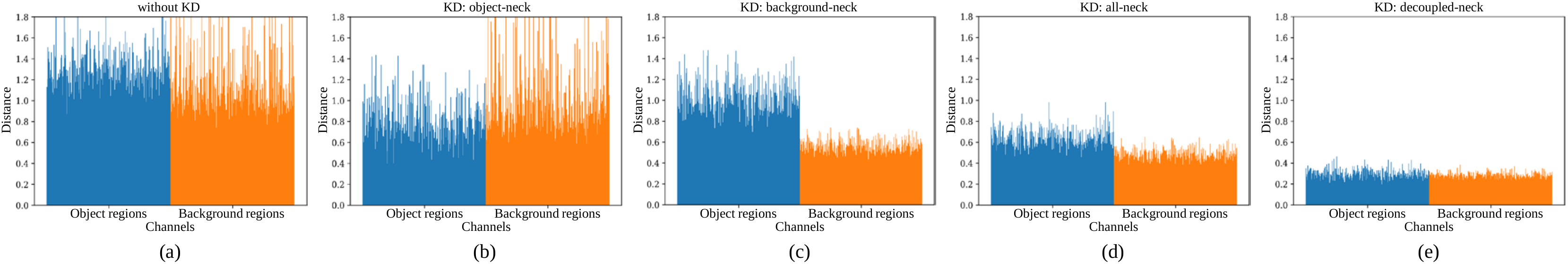}
	\vspace{-0.5cm}
	\caption{\small{Per-channel distance between teacher and student detector. (a) Student is trained without KD technique; (b) Student is distilled via object only regions in neck features; (c) Student is distilled via background only regions in neck features; (d) Student is distilled via all regions in neck features; (e) Student is distilled via decoupled regions (DeFeat) in neck features.}}
	\vspace{-0.4cm}
	\label{fig:vis}
\end{figure*}

\vspace{0.cm}
\noindent \textbf{Impact of positive and negative proposals in classification head.}
The evaluation of distillation on proposals are shown in Table \ref{table:proposal}, coefficients $\beta_{obj}$, $\beta_{bg}$, $T_{obj}$, $T_{bg}$ and $\lambda$ are illustrated in Equation~\ref{eq:kd-cls} and Equation~\ref{eq:previous-cls}. We can find that using positive only proposals in distillation can slightly boost the performance of student detector, but demands a smaller coefficient. Using negative only proposals in distillation achieves better performance compared to using positive only proposals. One main reason is that the numbers of positive and negative proposals are imbalanced, and the difficulty in optimizing these two types of proposals can be different. Figure~\ref{fig:loss_proposal} also indicates that the distillation loss of negative proposals drops faster than that of positive proposals. The ``sub-negative" in Table \ref{table:proposal} denotes that we randomly select samples with the same number of positive proposals from negative proposals, which decreases the mAP by 0.2\%, demonstrating that the distillation result is associated with the number of proposals. Our decoupled distillation improves the performance of previous method that treats all proposals equally from 38.2\% to 38.9\% mAP, which demonstrates the effectiveness of DeFeat.

\vspace{0.cm}
\noindent \textbf{Comparison of the shared proposals from teacher and student.}
Given a teacher detector and a student detector, the region proposals output by two models are inevitably different and consequently the student cannot be distilled directly. We analyze the performances of sharing teacher proposals with student and sharing student proposals with teacher. As can be seen in Table \ref{table:shared}, feeding the proposals from teacher into distillation performs better than feeding the proposals from student detector. One main reason is that due to the large amount of possible region proposals, the proposals from teacher detector would be more accurate and contain more intensive information for distillation.

\vspace{0.cm}
\noindent \textbf{Performance gain from object and background regions.} Figure \ref{fig:pie_exp} presents analyses on three randomly selected classes. Distillation via object regions and background regions both improve the number of correct detection significantly. Object regions can bring stronger localization ability (Loc) to the student, while background regions can effectively reduce the false positive rate (BG). 

\vspace{-0.cm}
\noindent \textbf{Distance between teacher and student.}
We calculate the per-channel distance between teacher and student to better understand why decoupled regions can attain higher performance compared to its counterparts. We randomly choose 20 images from COCO minival and calculate the absolute value of the discrepancy between the same region in neck features from teacher and student. As shown in Figure \ref{fig:vis}, the teacher and the student are ResNet152-FPN and ResNet50-FPN, respectively. X axis stands for each channel (256 in total) from the first layer of FPN. Y axis stands for the average distance of pixels belonging to object/background regions per channel. Distilling student detector via object regions in neck features can substantially narrow the distances in object regions, but fail to promote the background regions. Distilling student object via treating all regions equally can decrease the distance by a fair margin. Our proposed DeFeat could further minimize the distance to the largest extent for both object and background regions.

\section{Conclusion}
In this paper, we propose a simple yet efficient distillation method via decoupled features for object detection. We analyze and demonstrate the important roles of background regions during the distillation process. Based on ample observations, we introduce the DeFeat method in which the features are split into object and background parts at FPN level and RoI-aligned feature level, and distillation is applied on these two parts separately. DeFeat is general and can be easily used for both one-stage and two-stage detection frameworks. Extensive experiments validate the effectiveness of DeFeat by consistently outperforming other distillation techniques. \\
\textbf{Acknowledgement} Chang Xu was supported in part by the Australian Research Council under Projects DE180101438 and DP210101859. And we sincerely thank all reviewers and ACs for their valuable comments.

{\small
	\bibliographystyle{ieee_fullname}
	\bibliography{egbib}
}
	
\end{document}